\documentclass[letterpaper]{article} 
\usepackage{aaai2026}
\usepackage{times}  
\usepackage{helvet}  
\usepackage{courier}  
\usepackage[hyphens]{url}  
\usepackage{graphicx} 
\usepackage{natbib}  
\usepackage{caption} 
\usepackage{algorithm}
\usepackage{algorithmic}

\usepackage{newfloat}
\usepackage{listings}
\DeclareCaptionStyle{ruled}{labelfont=normalfont,labelsep=colon,strut=off} 
\floatstyle{ruled}
\newfloat{listing}{tb}{lst}{}
\floatname{listing}{Listing}
\usepackage{makecell}
\usepackage{subfigure}
\usepackage[graphicx]{realboxes}
\usepackage{multirow}
\usepackage{multicol}
\usepackage{booktabs}
\usepackage{amsfonts}
\usepackage{pifont}
\newcommand{\cmark}{\ding{51}} 
\newcommand{\xmark}{\ding{55}}           
\usepackage[most]{tcolorbox}
\tcbset{
    mypromptbox/.style={
    colback=gray!10,       
    colframe=gray!150,     
    arc=2.5mm,               
    boxrule=1.2pt,         
    left=4pt, 
    right=4pt, 
    top=4pt, 
    bottom=4pt,
    fonttitle=\bfseries,
    coltitle=black,
  }
}
\nocopyright

\title{Modality and Task Adaptation for Enhanced Zero-shot Composed Image Retrieval}
\author{
    Haiwen Li\textsuperscript{1}\quad Fei Su\textsuperscript{1,2}\quad Zhicheng Zhao\textsuperscript{1,2}
}

\affiliations{
    {\textsuperscript{1}Beijing University of Posts and Telecommunications} \\
    {\textsuperscript{2}Beijing Key Laboratory of Network System and Network Culture, China} \\
    \{lihaiwen, sufei, zhaozc\}@bupt.edu.cn

}

\usepackage{bibentry}

\begin{document}

\maketitle

\begin{abstract}
As a challenging vision-language task, Zero-Shot Composed Image Retrieval (ZS-CIR) is designed to retrieve target images using bi-modal (image+text) queries. Typical ZS-CIR methods employ an inversion network to generate pseudo-word tokens that effectively represent the input semantics. However, the inversion-based methods suffer from two inherent issues: First, the task discrepancy exists because inversion training and CIR inference involve different objectives. Second, the modality discrepancy arises from the input feature distribution mismatch between training and inference. To this end, we propose a lightweight post-hoc framework, consisting of two components: (1) A new text-anchored triplet construction pipeline leverages a large language model (LLM) to transform a standard image-text dataset into a triplet dataset, where a textual description serves as the target of each triplet. (2) The MoTa-Adapter, a novel parameter-efficient fine-tuning method, adapts the dual encoder to the CIR task using our constructed triplet data. Specifically, on the text side, multiple sets of learnable task prompts are integrated via a Mixture-of-Experts (MoE) layer to capture task-specific priors and handle different types of modifications. On the image side, MoTa-Adapter modulates the inversion network's input to better match the downstream text encoder. In addition, an entropy-based optimization strategy is proposed to assign greater weight to challenging samples, thus ensuring efficient adaptation. Experiments show that, with the incorporation of our proposed components, inversion-based methods achieve significant improvements, reaching state-of-the-art performance across four widely-used benchmarks. All data and code will be made publicly available.
\end{abstract}


\section{Introduction}
Composed Image Retrieval (CIR)~\cite{cir} has attracted growing interest in recent years, aiming to retrieve target images that align closely with a bi-modal query consisting of a reference image and a relative caption. Compared with traditional image retrieval~\cite{deepfusion}, CIR integrates visual and textual inputs to enable more flexible searches. With the availability of large-scale vision-language pretraining (VLP) models~\cite{clip,align,blip}, CIR has made substantial progress, facilitating its applications in various real-world scenarios, particularly in e-commerce and web search. 

Supervised CIR approaches~\cite{clip4cir,blip4cir,sprc} utilize labeled triplets $(I_r, T_c, I_t)$, where $I_r$ represents the reference image, $T_c$ denotes the relative caption, and $I_t$ is the target image. However, it is both labor-intensive and time-consuming to manually collect and annotate sufficient triplet data, which results in relatively small-scale datasets~\cite{cirr,fashioniq} and restricts the generalization capability. Instead of relying heavily on manually annotated triplets, Zero-Shot Composed Image Retrieval (ZS-CIR)~\cite{pic2word,serealcirco} has emerged as a scalable solution to overcome the limitations of supervised CIR. In comparison, ZS-CIR benefits from much larger and more diverse publicly available datasets~\cite{cc3m,laion} that encompass a broad range of domains and semantics, thereby equipping models with enhanced generalization ability.

\begin{figure}[t]
\centering
\includegraphics[width=0.96 \columnwidth]{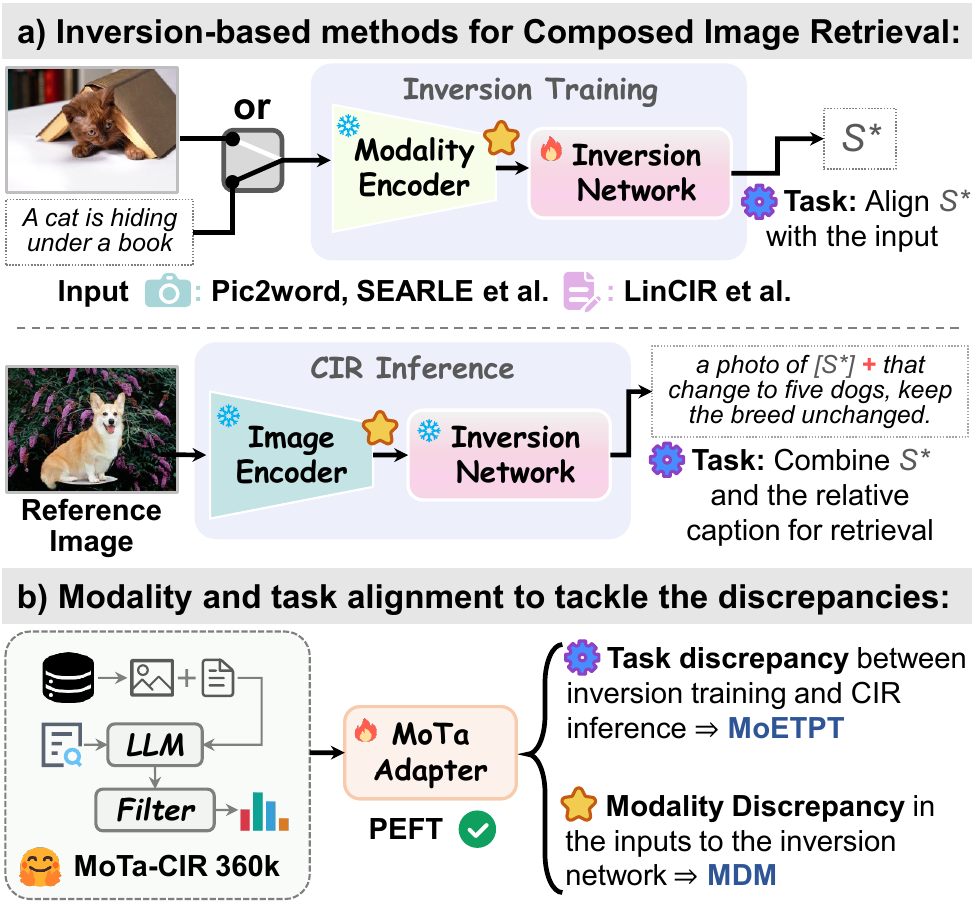}
\caption{\textbf{Motivation of our work.} a) Existing inversion-based approaches, encompassing both image-based methods such as Pic2Word~\cite{pic2word} and SEARLE~\cite{serealcirco}, and text-based methods like LinCIR~\cite{LinCIR}. b) Our proposed lightweight post-hoc framework aims at jointly addressing the task and modality discrepancies inherent in inversion-based CIR methods.}
\label{fig_intro}
\end{figure}

Mainstream ZS-CIR approaches are primarily based on textual inversion~\cite{personalizingword,textinv}, and can be categorized into image-based~\cite{pic2word,serealcirco} and text-based methods~\cite{LinCIR}. These methods train an inversion network that maps the features of an input image (or text) into a pseudo-word token, denoted as $S^*$. This token is subsequently incorporated into a prompt template (e.g., a photo of $S^*$) to represent the input semantics faithfully. However, as illustrated in Figure~\ref{fig_intro}(a), the inversion-based methods face two key issues between training and inference: (1) \textit{\textbf{Task discrepancy}}~\cite{rtd}. The task of inversion training is to align the pseudo-word token $S^*$ with the input image (or text), such that $S^*$ serves as a good representation of the input. In contrast, the task during CIR inference is to compose a query from $I_r$ and $T_c$ to retrieve target images $I_t$, which is not tackled during training. (2) \textit{\textbf{Modality discrepancy}}. This issue is particularly prominent in text-based inversion training, where the image modality is absent during training but required at inference. Although LinCIR~\cite{LinCIR} introduces random perturbations at the feature level to mitigate this discrepancy, we demonstrate that this strategy is far from sufficient.

To tackle the aforementioned two discrepancies and improve inversion-based approaches, we propose a lightweight post-hoc framework, as illustrated in Figure~\ref{fig_intro}(b). It introduces an additional adaptation stage that adopts a parameter-efficient tuning strategy on automatically generated text-anchored triplets, comprising two parts: (1) \textit{\textbf{MoTa-Adapter}}. The task discrepancy arises on the query-side, as the inversion training does not involve the integration of $I_r$ and $T_c$. Building on this, we perform task adaptation using the input in the form of ($I_r$, $T_c$) to learn the integration process on the query side. Specifically, we insert several learnable task prompts into the input of the text encoder and integrate them through a Mixture-of-Experts (MoE)~\cite{moe} mechanism to capture CIR task priors, and meanwhile handle diverse types of modifications associated with $T_c$. In addition, since the modality discrepancy exists between the the inversion network's inputs during inversion training and CIR inference. To deal with it, we propose Modality Distribution Modulation (MDM), which adaptively shifts and aligns the feature distribution fed into the inversion network to mitigate the modality discrepancy and better support the downstream text encoder. (2) \textit{\textbf{MoTa-CIR}}. Considering that task and modality adaptation require triplet data involving $I_r$ for training, we propose a pipeline to automatically generate text-anchored triplets in the form of $(I_r, T_c, T_t)$. Specifically, given a standard image-text dataset, we guide an LLM to expand each image-text pair $(I_r, T_r)$ into a triplet by generating $T_c$ and $T_t$ based on $T_r$. Compared with using $I_t$ as the target, $T_t$ offers lower computational cost and better training efficiency. Moreover, since CIR resembles a fuzzy retrieval task~\cite{fuzzy,region_fuzzy} and there may be many valid target images, using a textual target $T_t$ for training facilitates robust learning. Finally, after filtering, we obtain a high-quality text-anchored triplet dataset, MoTa-CIR, comprising approximately 360k samples. In summary, our contributions are fourfold: 
\begin{itemize}
\item We propose a lightweight post-hoc framework that effectively mitigates the two inherent and interrelated task and modality discrepancies in inversion-based methods.
\item We introduce the MoTa-Adapter, a novel parameter-efficient fine-tuning approach for CIR, which optimizes only a small set of learnable task prompts on the text side and a modulation layer on the image side, addressing the two discrepancies simultaneously.
\item To facilitate task and modality adaptation, we propose a scalable pipeline for the automatic construction of text-anchored triplets, resulting in a diverse and high-quality dataset, MoTa-CIR. Additionally, we introduce a novel entropy-based loss weighting strategy for efficient training on the text-anchored triplets.
\item When our proposed modules are integrated into the existing inversion-based methods, the performance is significantly enhanced across four widely-used benchmarks.
\end{itemize}

\section{Related work}
\label{sec_related}
\noindent\textbf{Composed Image Retrieval (CIR)} is primarily evaluated in the fashion~\cite{fashioniq} and real-world~\cite{cirr,serealcirco} domains. Mainstream supervised approaches~\cite{clip4cir,blip4cir,sprc} leverage the cross-modal alignment capabilities of the VLP models and adopt either early or late fusion to integrate the two modalities in composed queries. Recent work explores zero-shot CIR, with textual inversion~\cite{textinv,personalizingword} emerging as a key technique. Representative methods, including image-based approaches such as Pic2Word~\cite{pic2word} and SEARLE~\cite{serealcirco}, as well as the text-based methods like LinCIR~\cite{LinCIR}, train an inversion network to generate a pseudo-word token that effectively captures and represents the input semantics. Additionally, some attempts~\cite{contextI2W,rtd,cig,predicir} have improved upon inversion training, in which most related to ours is RTD~\cite{rtd}. Unlike RTD, which only fine-tunes the text encoder to reduce the task discrepancy, our approach introduces a lightweight adapter that enables multi-modal adaptation of the dual encoder, effectively addressing the inherent limitations of inversion-based methods. Other works explore automatic CIR triplet construction~\cite{covr,case,compodiff} and training-free approaches based on LLM reasoning~\cite{cirevl,ldre}. While both have shown promising results, they are limited by the quality of constructed triplets and model complexity, respectively.

\textbf{Parameter-Efficient Fine-Tuning (PEFT)} aims to adapt large-scale pretrained VLP models by updating only a small amount of task-specific parameters, reducing overhead while preserving performance. Methods such as Adapter Tuning~\cite{adaptertuning}, Prompt Tuning~\cite{prompttuning}, and LoRA~\cite{lora} have shown strong results in NLP. In the vision-language domain, approaches like CoOp~\cite{coop}, CoCoOp~\cite{cocoop} and Maple~\cite{maple} adapt CLIP using context prompts or prompt modules for better generalization and robustness. Our work builds on this line by introducing a PEFT-based strategy tailored for CIR, jointly addressing both task and modality discrepancies.

\begin{figure*}[t]
\centering
\includegraphics[width=1\textwidth]{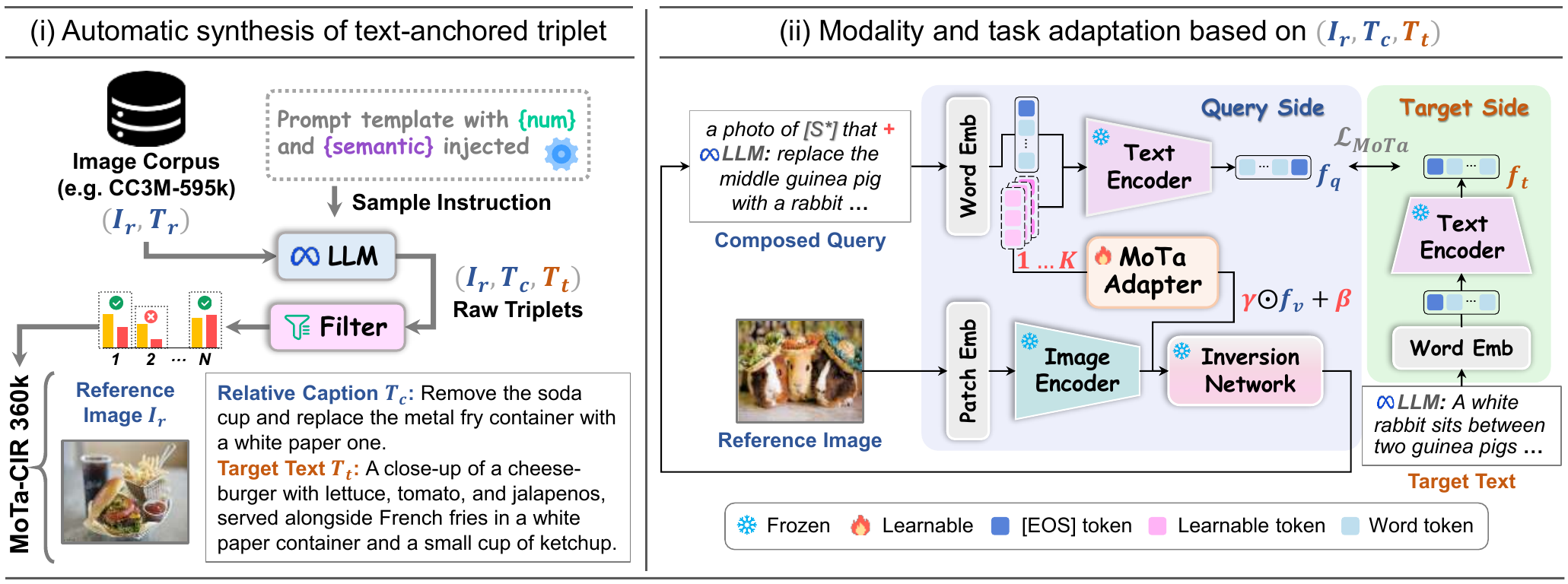}
\caption{\textbf{Overview of our proposed framework.} (i) A standard image-text dataset containing $(I_r,T_r)$ is first selected. An LLM is guided to generate diverse $(T_c,T_t)$ based on $T_r$. Subsequently, the raw triplets $(I_r,T_c,T_t)$ are scored and filtered by an MLLM to form the MoTa-CIR dataset. (ii) Based on the text-anchored triplets constructed in (i), the lightweight MoTa-Adapter is incorporated to explicitly reduce task and modality discrepancies.
It operates on both the text and image sides, enhancing the model’s capacity to interpret composed queries, thereby improving overall CIR performance.}
\label{fig_main}
\end{figure*}

\section{Methodology}
\label{sec_method}
In this section, we introduce our proposed approach in detail. ~\ref{subsec_gen_triplet} presents the pipeline for automatically constructing text-anchored triplets, including diversified triplet expansion and data filtering. ~\ref{subsec_mota} describes the modality and task adaptation, including the design of the MoTa-Adapter and its role in model optimization and the inference workflow.

\subsection{Automatic Synthesis of Text-anchored Triplet}
\label{subsec_gen_triplet}
As supervised CIR relies on manually annotated triplets, several studies have proposed using cheaply and automatically collected triplets for training. Among these, the most closely related to our work is RTD~\cite{rtd}, but it adopts text-only triplets in the form of $(T_r, T_c, T_t)$. 


\textbf{Diversified Triplet Expansion.} Unlike RTD, which attributes the task discrepancy only to the text encoder, we argue that the discrepancy lies in the dual-encoder structure. In particular, the integration of the reference image $I_r$ and the relative caption $T_c$ on the query side is essential for effectively bridging the task gap. To this end, one or several publicly available image-text datasets are selected, and their union is denoted as $\mathcal{D} = \{{I_r^i, T_r^i}\}_{i=1}^{N}$. For each $T_r^i$, a template $\mathcal P(num,semantic)$, as illustrated below, guides an LLM to generate a relative caption $T_c^i$ and a target text $T_t^i$, resulting in a triplet of the form $(I_r^i, T_c^i, T_t^i)$.

\begin{tcolorbox}[mypromptbox, halign=left]
Given the reference caption \{$\boldsymbol{T_r}$\}, please generate an editing instruction involving \{\textbf{\textit{num}}\} modifications on the \{\textbf{\textit{semantic}}\} aspect, along with a target text.
\end{tcolorbox}

Note that $num$ is sampled from 1 to 3, and $semantic$ is drawn from a predefined set covering operations such as attribute modification, substitution, and removal. This guarantees that $T_c$ includes varying types and quantities of modifications, thereby achieving satisfactory semantic diversity.

\textbf{Data Filtering.} To enhance the overall quality of the generated triplets, we use a multi-modal large language model (MLLM) as a filtering module to score each triplet on two dimensions, ranging from 1 to 10: (1) Triplet alignment: how well the composition of $I_r$ and $T_c$ semantically aligns with $T_t$, and (2) Variety: whether $T_t$ involves meaningful and diverse modifications relative to $T_r$. For each dimension, the bottom 30\% of triplets are discarded based on their scores.

Using this pipeline, we construct a diverse and high-quality text-anchored triplet dataset, MoTa-CIR-360k. This dataset is used to support the modality and task adaptation. More details and the prompts used for the LLM and MLLM are provided in the Appendix A.2.

\subsection{Modality and Task Adaptation} 
\label{subsec_mota}
A brief introduction of CLIP and inversion-based methods are first given, and then our proposed MoTa-Adapter and entropy-based optimization are introduced in detail.

\textbf{Preliminary.} CIR models are mainly built on CLIP~\cite{clip}, which is trained on massive image-text pairs and exhibits strong cross-modal alignment capabilities. On the text side, CLIP employs Byte-Pair Encoding (BPE)~\cite{bpe} to tokenize the input into a sequence of tokens $E = \{{e_i} \in \mathbb{R}^{d_e}\}_{i=1}^M$, which is then fed into the text encoder $\mathcal{T}$ to obtain the text feature $f_t=\mathcal T(E) \in \mathbb{R}^{d}$. On the image side, it follows a similar process by extracting image features $f_v \in \mathbb{R}^d$ using the image encoder $\mathcal{I}$. The resulting features $f_t$ and $f_v$ are aligned in a joint embedding space via a contrastive loss~\cite{moco}.

Inversion-based approaches utilize an inversion network $\phi$ to project an input image (or text) feature $f\in \mathbb R^{d}$ into the textual embedding space, producing a pseudo-word token $e_{s^*}=\phi(f) \in \mathbb{R}^{d_e}$ that is inserted into the prompt “a photo of $S^*$”. After tokenization, the resulting sequence is fed into $\mathcal{T}$ to obtain the inverse feature $f^*\in \mathbb R^{d}$, which is trained to be close to the original feature $f$ by contrastive learning. This allows “a photo of $S^*$” to serve as an effective representation of the input image (or text).

\textbf{MoTa-Adapter.} However, we notice that the lack of integration between $I_r$ and $T_c$ during inversion training is the primary cause of the task and modality discrepancies. Therefore, it is necessary to adapt the dual encoder.

The modality discrepancy arises because of the misalignment in the feature distributions input to the $\phi$ network during inversion training and CIR inference. Although VLMs like CLIP have performed cross-modal alignment, internal covariance and scale differences still exist within the distributions of visual and textual modalities. Therefore, as shown in Figure~\ref{fig_main}(ii), MoTa-Adapter incorporates Modality Distribution Modulation (MDM) on the image side, where a modulation layer after the image encoder $\mathcal{I}$ is built to adaptively shift the distribution of $f_v$ using a learnable scale parameter $\gamma \in \mathbb{R}^{d}$ and a learnable shift parameter $\beta \in \mathbb{R}^{d}$:
\begin{equation}
\tilde{f_v} = \gamma \odot f_v+\beta,
\end{equation}
\noindent where $\odot$ represents the inner product, and the modulated image features are then mapped to the text side using the inversion network, i.e., $e_{s^*} = \phi(\tilde{f_v}) \in \mathbb{R}^{d_e}$, in which $d_e$ is the dimension of the textual embedding space. MDM not only reduces the modality discrepancy but also integrates the internal information of the image features to better resolve task discrepancy. In other words, the two discrepancies are interrelated and require simultaneous adaptation.

Meanwhile, the task adaptation should enable the text encoder $\mathcal T$ to understand “a photo of $S^*$ that $T_c$” rather than just “a photo of $S^*$”. Building on this, our MoTa-Adapter introduces Mixture-of-Experts Task Prompt Tuning (MoE-TPT) on the text side, which inserts $K$ experts, each containing $N$ learnable task prompts $P_k=\{p_i\in\mathbb{R}^{d_e}\}_{i=1}^N, k=1,2,...,K$, into the input token sequence of $\mathcal T$. The $K$ experts are fused through a routing function $\mathcal R$ and are conditioned on the pseudo-word representation $e_{s^*}$:
\begin{equation}
\begin{aligned}
    & \mathcal R(x)_k = \text{Softmax}(\mathcal{W}x)_k \\
    & P = \Sigma{_{k=1}^{K}} P_k\cdot \mathcal R(e_{s^*})_k,
\end{aligned}
\end{equation}
\noindent where $\mathcal W \in \mathbb{R}^{d_e \times K}$ is a linear layer, generating the weights of $K$ experts based on $e_{s^*}$. Subsequently, $P$ is concatenated with other textual tokens and input into $\mathcal{T}$ to obtain the composed feature $\tilde{f}_c=\mathcal T([P,E])$, which is then optimized to inject the task-specific priors into the CIR model. The MoE structure provides two main advantages for the CIR task: (1) \textit{\textbf{Task Specialization}}, each expert focus on different types of modifications in the relative captions, such as attribute manipulation, substitution, or removal. (2) \textit{\textbf{Sample Customization}}, the routing function $\mathcal{R}$ generates weights based on the pseudo-word representation $e_{s^*}$, assigning the optimal mixture of experts for each sample.

\textbf{Entropy-based Optimization.} Given that the inversion-trained models already possess a certain level of CIR capabilities, we propose an entropy-based optimization strategy that assigns greater attention to challenging samples. Specifically, for each composed feature $f_c^i$ produced by the base model (w/o MoTa-Adapter), the cosine similarities with all target text features $\{f_t^j\}_{\mathcal B}$ in a batch are computed. Based on the predicted probabilities, the entropy can be quantified:
\begin{equation}
\begin{aligned}
p_{c2t}^{(i,j)}&=\frac{e^{sim(f_c^i,f_t^j)}}{\Sigma_{j\in \mathcal B}e^{sim(f_c^i,f_t^j)}}\\
H_{c2t}^i &= -\sum_j p^{(i,j)}_{c2t} \log p^{(i,j)}_{c2t},
\end{aligned}
\end{equation}
\noindent where $sim(\cdot,\cdot)$ denotes the cosine similarity. A larger entropy $H^i_{c2t}$ indicates higher predictive uncertainty of the base model for $f_c^i$, suggesting it should be assigned greater weight during the adaptation. We normalize and apply exponential smoothing to compute the sample-wise weight $w^i_{c2t}$:
\begin{equation}
\hat{H}^i_{c2t} = \frac{H^i_{c2t}}{\log |\mathcal B|} \quad w^i_{c2t} = e^{\beta \hat{H}^i_{c2t}},
\end{equation}
\noindent the division is based on Jensen’s inequality, yielding $H^i_{c2t} \leq \log |\mathcal B|$, where $|\mathcal B|$ denotes the batch size. $\beta$ is a hyperparameter controlling the sharpness of the exponential smoothing function. Similarly, the weight $w^i_{t2c}$ for each target text feature $f_t^i$ can be obtained. Finally, the tuned model (w/ MoTa-Adapter) generates $\tilde{f}_c$ and $\tilde{f}_t$ that are optimized using a weighted contrastive loss, i.e. $\mathcal L_{MoTa}$, in which $\tau$ is a learnable temperature parameter,
\begin{equation}
\begin{aligned}
    \mathcal L_{MoTa}(\mathcal B)=&-\sum_{i=1}^{|\mathcal B|} w^i_{c2t}\cdot \mathrm{log}[\frac{e^{sim(\tilde{f}_c^i, \tilde{f}_t^i)/\tau}}{\Sigma_{j\in \mathcal B} e^{sim(\tilde{f}_c^i,\tilde{f}_t^j)/\tau}}]\\
    & -\sum_{i=1}^{|\mathcal B|} w^i_{t2c}\cdot \mathrm{log}[\frac{e^{sim(\tilde{f}_t^i, \tilde{f}_c^i)/\tau}}{\Sigma_{j\in \mathcal B} e^{sim(\tilde{f}_t^i, \tilde{f}_c^j)/\tau}}].
\end{aligned}
\end{equation}

The lower bound occurs when all samples have identical predicted entropy, resulting in equal weights. In this case, the loss degenerates into the standard contrastive loss.

\subsection{Inference Workflow}
During inference, the image encoder is used to extract features of an image gallery, resulting in a set $\mathcal{S}=\{f_v^i\}_{i=1}^N$. After that, given an input composed query $(I_r, T_c)$, the composed feature $\tilde f_{c}$ is obtained following the procedure described in Section~\ref{subsec_mota}. The cosine similarity between $\tilde f_{c}$ and each target image feature in $\mathcal{S}$ is computed, with the top-K most similar items returned as the retrieval results. 


\begin{table*}[t]
\small
\centering
\begin{tabular}{l | c | c c | c c c | c c | c}
\toprule
\multirow{3}{*}{\textbf{Method}} & \multirow{3}{*}{\textbf{Backbone}} & \multicolumn{8}{c}{\textbf{Zero-shot Composed Image Retrieval}} \\
& & \multicolumn{2}{c}{FashionIQ} & \multicolumn{3}{c}{CIRR} & \multicolumn{2}{c}{CIRCO} & \multicolumn{1}{c}{GeneCIS} \\
\cmidrule(lr){3-4} \cmidrule(lr){5-7} \cmidrule(lr){8-9} \cmidrule(lr){10-10}
& & R@10 & R@50 & R@1 & R@5 & R@10 & mAP@5 & mAP@10 & R@1 \\
\midrule
CoVR-BLIP~\cite{covr} & BLIP & 27.70 & 44.63 & 38.48 & 66.70 & 77.25 & 21.43 & 22.33 & - \\
CASE~\cite{case} & BERT+ViT & - & - & 35.40 & 65.78 & 78.53 & - & - & - \\
CIReVL$^\dagger$~\cite{cirevl} & CLIP-G & 32.19 & 34.65 & 64.29 & 67.95 & 75.06 & 26.77 & 27.59 & 17.4 \\
CompoDiff$^\dagger$~\cite{compodiff} & CLIP-G & 39.02 & 51.71 & 26.71 & 55.14 & 74.52 & 15.33 & 17.71 & 15.5 \\
\midrule
\multicolumn{10}{c}{\textit{Comparison with methods based on textual inversion}} \\
\midrule
ContextI2W~\cite{contextI2W} & \multirow{3}{*}{CLIP-L} & 27.80 & 48.90 & 25.60 & 55.10 & 68.50 & 13.00 & 14.60 & 12.7 \\
KEDs~\cite{ked} &  & 26.80 & 47.90 & 26.40 & 54.80 & 67.20 & - & - & - \\
PrediCIR~\cite{predicir} &  & 30.10 & 52.30 & 27.20 & 57.00 & 70.20 & 15.70 & 17.10 & 16.6 \\
\midrule
Image2Sentence~\cite{image2sentence} & BLIP & 29.79 & 49.19 & 29.68 & 58.72 & 70.79 & 9.67 & 10.32 & - \\
Slerp+TAT~\cite{slerp} & BLIP & 32.77 & 53.32 & 33.98 & 61.74 & 72.70 & 17.84 & 18.44 & - \\
PrediCIR~\cite{predicir} & CLIP-G & 47.20 & 67.80 & 37.00 & 66.10 & 77.90 & 23.70 & 24.60 & 18.7 \\
\midrule
Pic2Word~\cite{pic2word} & \multirow{12}{*}{CLIP-L} & 24.70 & 43.70 & 23.90 & 51.70 & 65.30 & 8.72 & 9.51 & 11.2 \\
+ CIG$^\dagger$~\cite{cig} &  & 25.16 & 44.85 & 24.63 & 52.75 & 65.28 & - & - & - \\
+ RTD~\cite{rtd} &  & 27.59 & \textbf{48.90} & 27.86 & 56.24 & 68.48 & 9.13 & 9.63 & 11.9 \\
\textbf{+ MoTa-Adapter (Ours)} &  & \textbf{27.73} & 48.15 & \textbf{28.06} & \textbf{57.22} & \textbf{69.37} & \textbf{9.79} & \textbf{10.18} & \textbf{12.1} \\
\cmidrule(lr){1-1} \cmidrule(lr){3-10}
SEARLE~\cite{serealcirco} &  & 25.56 & 46.23 & 24.24 & 52.48 & 66.29 & 11.68 & 12.73 & 12.3 \\
+ CIG$^\dagger$~\cite{cig} &  & 25.66 & 46.50 & 26.72 & 55.52 & 68.10 & 12.84 & 13.64 & - \\
+ RTD~\cite{rtd} &  & \textbf{29.34} & \textbf{50.73} & 26.63 & 56.17 & 68.96 & 16.53 & 17.89 & 12.4 \\
\textbf{+ MoTa-Adapter (Ours)} &  & 27.78 & 48.51 & \textbf{28.19} & \textbf{57.95} & \textbf{69.98} & \textbf{16.77} & \textbf{17.94} & \textbf{14.7} \\
\cmidrule(lr){1-1} \cmidrule(lr){3-10}
LinCIR~\cite{LinCIR} &  & 26.28 & 46.49 & 25.04 & 53.25 & 66.68 & 12.59 & 13.58 & 12.2 \\
+ CIG$^\dagger$~\cite{cig} &  & 26.60 & 47.22 & 26.17 & 54.94 & 67.64 & 12.84 & 13.77 & 12.2 \\
+ RTD~\cite{rtd} &  & \textbf{30.24} & \textbf{51.08} & 26.63 & 56.17 & 68.96 & 17.11 & 18.11 & 13.2 \\
\textbf{+ MoTa-Adapter (Ours)} & & 28.76 & 49.53 & \textbf{28.02} & \textbf{58.62} & \textbf{71.06} & \textbf{18.22} & \textbf{19.46} & \textbf{16.7} \\
\midrule
LinCIR~\cite{LinCIR} & \multirow{4}{*}{CLIP-G} & 45.11 & 65.69 & 35.25 & 64.72 & 76.05 & 19.71 & 21.01 & 13.6 \\
+ CIG$^\dagger$~\cite{cig} &  & 45.80 & 66.35 & 35.47 & 66.00 & 76.89 & 20.62 & 21.82 & 13.8 \\
+ RTD~\cite{rtd} &  & 46.21 & 67.26 & 36.31 & 67.47 & 78.31 & 21.08 & 22.29 & - \\
\textbf{+ MoTa-Adapter (Ours)} & & \textbf{47.06} & \textbf{67.37} & \textbf{38.39} & \textbf{69.47} & \textbf{80.05} & \textbf{25.82} & \textbf{27.06} & \textbf{19.1} \\
\bottomrule
\end{tabular}
\caption{\textbf{Performance comparison with existing zero-shot CIR methods.} The best results are marked in bold. $\dagger$ indicates the method that, in addition to using a retrieval backbone, incorporates complex auxiliary network structures such as Diffusion Models (DMs) or Large Language Models (LLMs), which results in lower inference efficiency.}
\label{tbl:zs-sota}
\end{table*}

\begin{table}[t]
\centering
\small
\begin{tabular}{l | c | c | c}
\toprule
\multirow{2}{*}{\textbf{Method}} & CIRR & FashionIQ & CIRCO \\ 
 & MeanR & MeanR & mAP@5 \\
\midrule
LinCIR~\cite{LinCIR} & 58.67 & 55.40 & 19.71 \\
+ Full Fine-tuning & 59.76 & 55.97 & 22.43 \\
+ CoOp~\cite{coop} & 61.48 & 56.55 & 24.89 \\
+ CoCoOp~\cite{cocoop} & 60.19 & 57.04 & 21.42 \\
+ Maple~\cite{maple} & 61.50 & 56.87 & 25.21 \\
\textbf{+ MoTa-Adapter (Ours)} & \textbf{62.64} & \textbf{57.22} & \textbf{25.82} \\
\bottomrule
\end{tabular}
\caption{\textbf{Results with different tuning strategies.} Experiments are conducted based on the CLIP-G backbone.}
\label{tbl:peft}
\end{table}

\begin{table}[t]
\centering
\small
\begin{tabular}{l | c | c}
\toprule
\multirow{2}{*}{\textbf{Dataset}} & CIRR & FashionIQ \\ 
 & MeanR & MeanR \\
\midrule
LinCIR~\cite{LinCIR} & 58.67 & 55.40 \\
+ST18M~\cite{compodiff} & 50.18 & 52.76 \\
+LaSCo~\cite{case} & 53.74 & 55.80 \\
+WebVid-CoVR~\cite{covr} & 58.25 & 56.18 \\
\textbf{+MoTa-CIR (Ours)} & \textbf{62.64} & \textbf{57.22} \\
\bottomrule
\end{tabular}
\caption{\textbf{Results with different datasets.} Experiments are conducted based on the MoTa-Adapter, using CLIP-G.}
\label{tbl:dataset}
\end{table}

\begin{table}[t]
\centering
\small
\begin{tabular}{l | c | c | c}
\toprule
\multirow{2}{*}{\textbf{Dataset}} & \multirow{2}{*}{\textbf{Filter}} & CIRR & FashionIQ \\ 
 & & MeanR & MeanR \\
\midrule
\multirow{2}{*}{LLaMA3-8B~\cite{llama3}} & \xmark & 62.26 & 56.81 \\
 & \cmark & \textbf{62.64} & 57.22 \\
 \midrule
\multirow{2}{*}{Qwen2.5-7B~\cite{qwen2.5}} & \xmark & 62.07 & 56.68 \\
 & \cmark & 62.30 & 56.87 \\
 \midrule
\multirow{2}{*}{Qwen2.5-32B~\cite{qwen2.5}} & \xmark & 62.03 & 56.74 \\
 & \cmark & 62.45 & \textbf{57.26} \\
\bottomrule
\end{tabular}
\caption{\textbf{Results with different LLMs for triplet expansion.} Experiments are conducted based on LinCIR+MoTa-Adapter, using CLIP-G as the backbone.}
\label{tbl:llm}
\end{table}

\section{Experiments}
\subsection{Experimental Setup}
\textbf{Evaluation benchmarks.} FashionIQ~\cite{fashioniq} simulates a realistic online shopping environment, featuring images in fashion domain. It comprises 30,134 triplets derived from 77,684 images. CIRR~\cite{cirr} is the first open-domain dataset in CIR, collecting 21,552 real-life images, with human-annotated relative captions. CIRCO~\cite{serealcirco} utilizes real-world images from the COCO dataset~\cite{coco} to develop a benchmark tailored for ZS-CIR with multiple ground truths. GeneCIS~\cite{genecis} assesses the models' ability to adapt to various notions of visual similarity given different text prompts. More details are provided in Appendix A.1.

\textbf{Evaluation metrics.} Evaluation of the model performance primarily employs the Rank-K (R@K) metric, which measures the probability of finding at least one target image within the top-K candidates. Specifically for CIRCO, the mean Average Precision (mAP) is the main criterion. Higher values in R@K and mAP indicate better performance.

\textbf{Implementation details.} (1) For the triplet expansion in Section~\ref{subsec_gen_triplet}, we select LLaMA3-8B~\cite{llama3} and convert the CC3M-595k~\cite{llava} dataset into a text-anchored triplet dataset. We then apply Qwen2.5-VL-32B~\cite{qwen2.5vl} for filtering, resulting in our MoTa-CIR-360k. The prompts we used, along with additional details can be found in Appendix A.2. (2) During the adaptation stage, for the MoTa-Adapter, we select $K=4$ experts, each associated with $N=8$ task prompts. We employ the AdamW optimizer~\cite{adamw} with a learning rate of $2e-3$, a weight decay of $0.01$, and a batch size of $256$. The model is trained for $1,000$ steps including $100$ warm-up steps on a single NVIDIA A100 GPU. More implementation details are provided in Appendix B.1.

\begin{table}[t]
\centering
\small
\begin{tabular}{l | c | c | c}
\toprule
\multirow{2}{*}{\textbf{Dataset}} & CIRR & FashionIQ & CIRCO \\ 
 & MeanR & MeanR & mAP@5 \\
\midrule
LinCIR~\cite{LinCIR} & 58.67 & 55.40 & 19.71 \\
w/ TPT (CoOp) & 61.48 & 56.55 & 24.89 \\
w/ MoE-TPT & 61.76 & 56.88 & 25.09 \\
w/ MDM & 59.80 & 56.32 & 22.16 \\
w/ MoE-TPT+MDM & 62.12 & 56.96 & 25.43 \\
\textbf{w/ MoE-TPT+MDM+EBO} & \textbf{62.64} & \textbf{57.22} & \textbf{25.82} \\
\bottomrule
\end{tabular}
\caption{\textbf{Ablation study.} Experiments are conducted based on the CLIP-G backbone.}
\label{tbl:abl}
\end{table}

\begin{figure}[t]
\centering
\includegraphics[width=1\columnwidth]{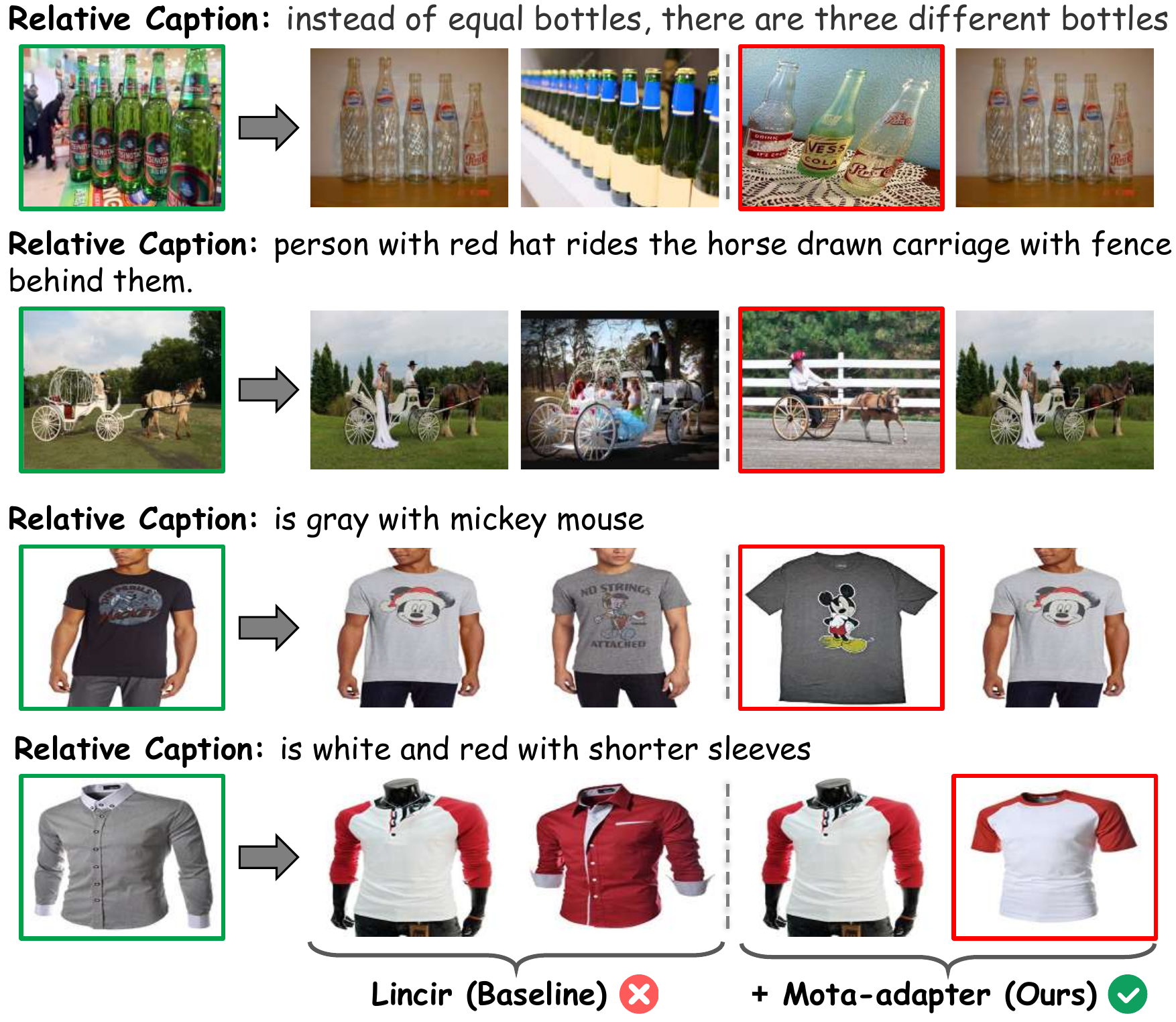}
\caption{\textbf{Qualitative results on CIRR and FashionIQ.} The reference image and its corresponding target image are highlighted with green and red outline.}
\label{fig:vis}
\end{figure}

\subsection{Quantitative Results} 
\noindent\textbf{Comparison with State-of-the-Art Methods.} As shown in Table~\ref{tbl:zs-sota}, we conduct experiments based on three baselines: two image-based methods, i.e., Pic2Word and SEARLE, and a text-based method LinCIR. (1) \textit{\textbf{Consistent Improvements.}} MoTa-Adapter consistently improves model performance across all evaluation metrics, no matter which baseline used. This demonstrates its effectiveness in addressing the inherent limitations of inversion-based methods. (2) \textit{\textbf{Essence of Discrepancy.}} The task discrepancy lies in the model's inability to effectively integrate $I_r$ and $T_c$ on the query side. RTD fine-tunes only the text encoder on pure text triplets, thus failing to fully resolve this issue. In contrast, our MoTa-Adapter adapts the dual encoder, reducing both task and modality discrepancies at a deeper level. This is further supported by empirical results, where MoTa-Adapter significantly outperforms both RTD and CIG on CIRR, CIRCO, and GeneCIS. On FashionIQ, MoTa-Adapter slightly underperforms RTD, which can be attributed to the domain gap between the real-world text-anchored triplets we construct and the fashion-related data in FashionIQ. (3) \textit{\textbf{Efficient Optimization.}} MoTa-Adapter is more efficient and resource-friendly. For example, CIG introduces T2I models~\cite{ldm}, which incur substantial computational costs during both training and inference. Training-free approaches such as CIReVL leverage LLMs for reasoning, leading to high resource consumption and slow inference speed. In contrast, our method introduces minimal learnable parameters and completes training in just 2 hours, making it both efficient and convenient. (4) \textit{\textbf{Inference Efficiency.}} We evaluate the per-query inference time on a single A100 GPU. Built on CLIP-G, LinCIR requires \textbf{0.011s}, and with the MoTa-Adapter, the time increases to \textbf{0.013s}. In contrast, CIReVL takes over \textbf{1s}, indicating that MoTa-Adapter introduces minimal inference overhead.

\textbf{Comparison with Different Tuning Strategies.} Table~\ref{tbl:peft} compares several tuning strategies on the MoTa-CIR dataset. (1) PEFT-based methods outperform full fine-tuning while being more lightweight. For example, MoTa-Adapter uses fewer than \textbf{1M} trainable parameters, whereas full fine-tuning updates the entire CLIP dual encoder (\textbf{441M}). (2) Multi-modal fine-tuning (Maple and MoTa-Adapter) is more effective than uni-modal approaches (CoOp and CoCoOp), and our MoTa-Adapter, specifically designed for the CIR task, achieves the best results.

\textbf{Comparison with Different Training Datasets.} Table~\ref{tbl:dataset} shows that only our MoTa-CIR consistently improves performance, while the other three datasets lead to either degradation or no clear gain. This suggests that MoTa-Adapter, as a PEFT method, is sensitive to data quality. Our pipeline generates high-quality, well-aligned text-anchored triplets, whereas the others may introduce noise due to low-quality images, repetitive relative captions, and poor alignment.

\begin{figure*}[t]
\centering
\subfigure[Impact of NTPs, NEs and $\beta$.]{
\includegraphics[width=0.31\textwidth]{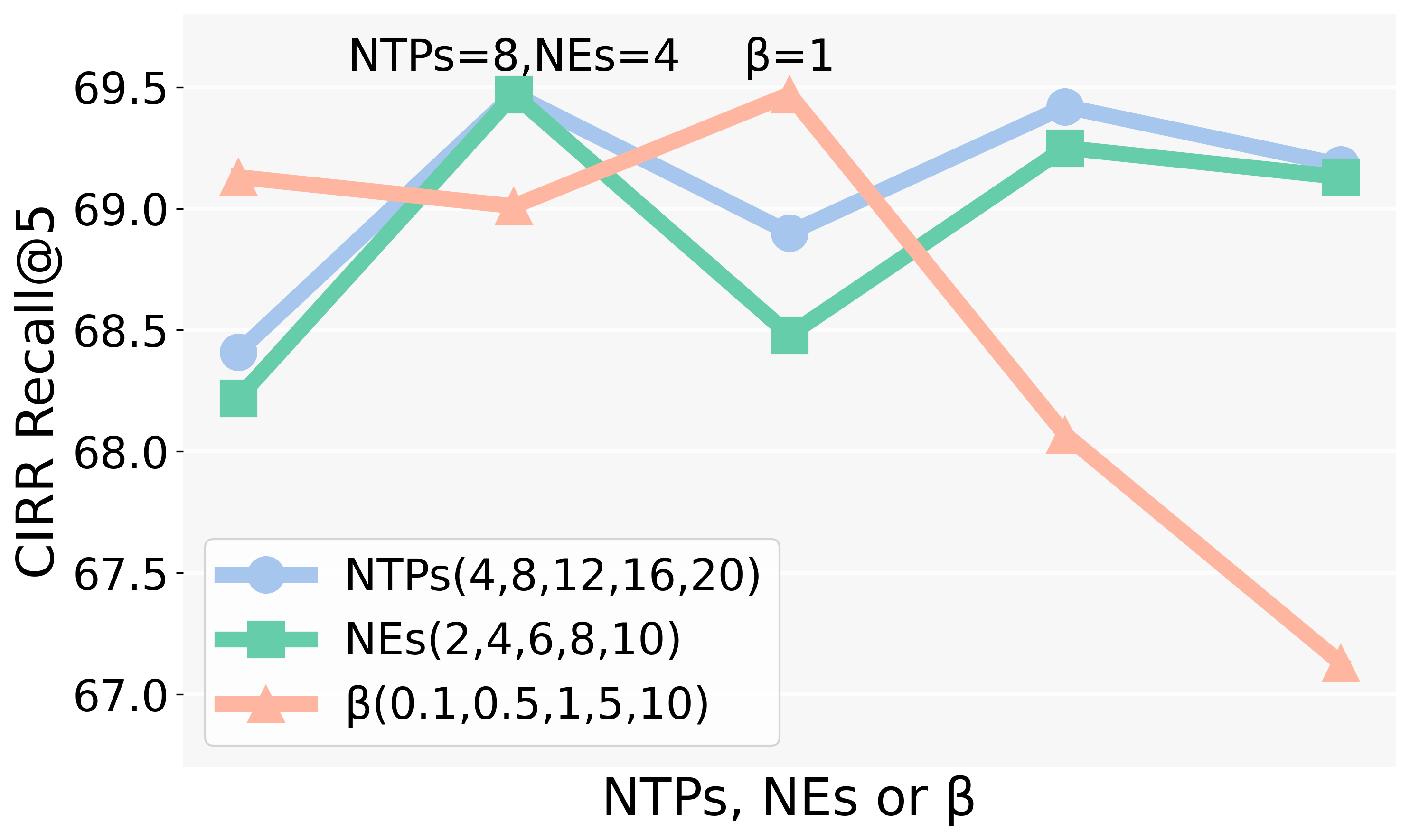} 
\label{fig:ablcurve}
}
\subfigure[Data efficiency of MoTa-Adapter.]{
\includegraphics[width=0.36\textwidth]{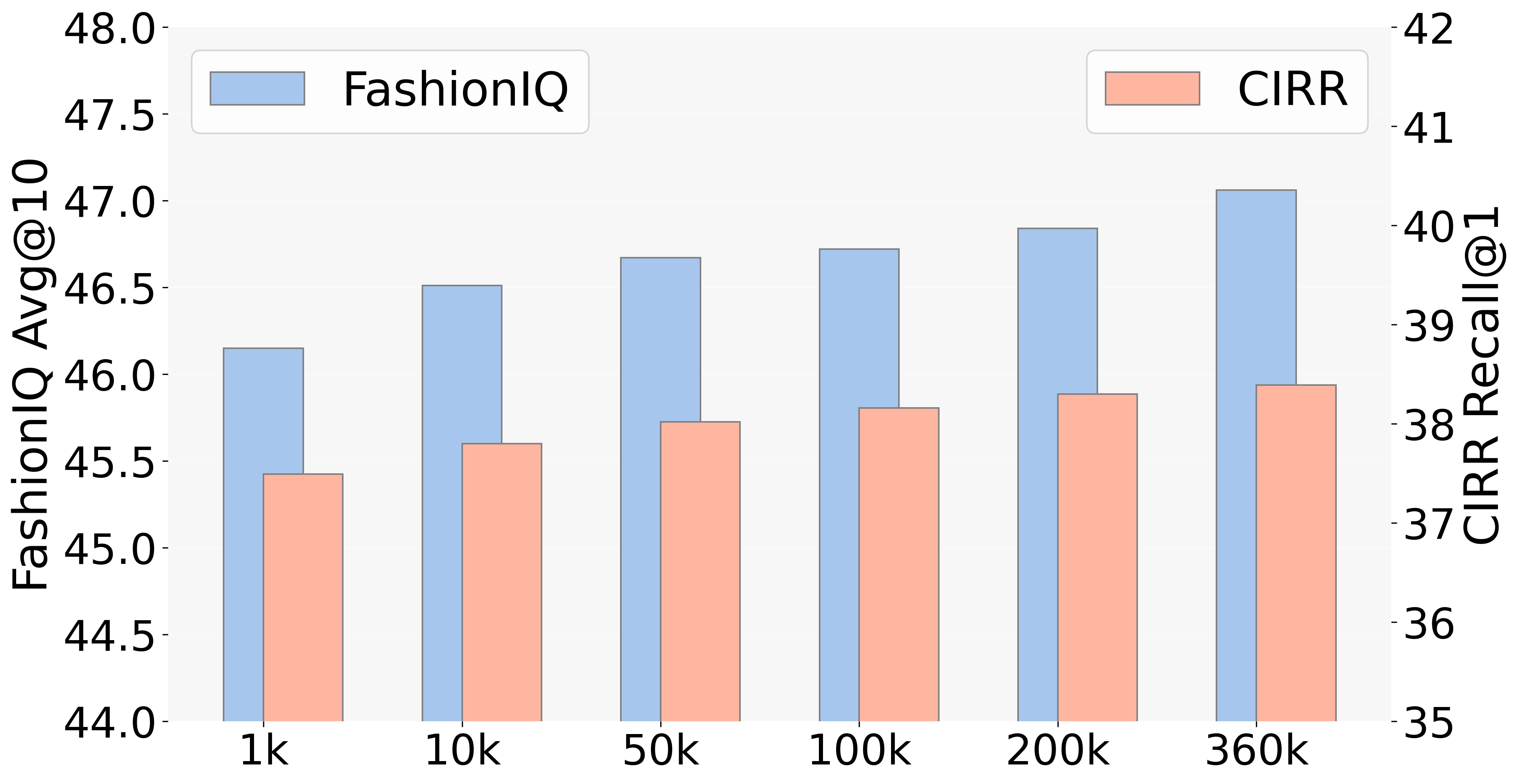}
\label{fig:quantity}
}
\subfigure[Visualization of training loss.]{
\includegraphics[width=0.29\textwidth]{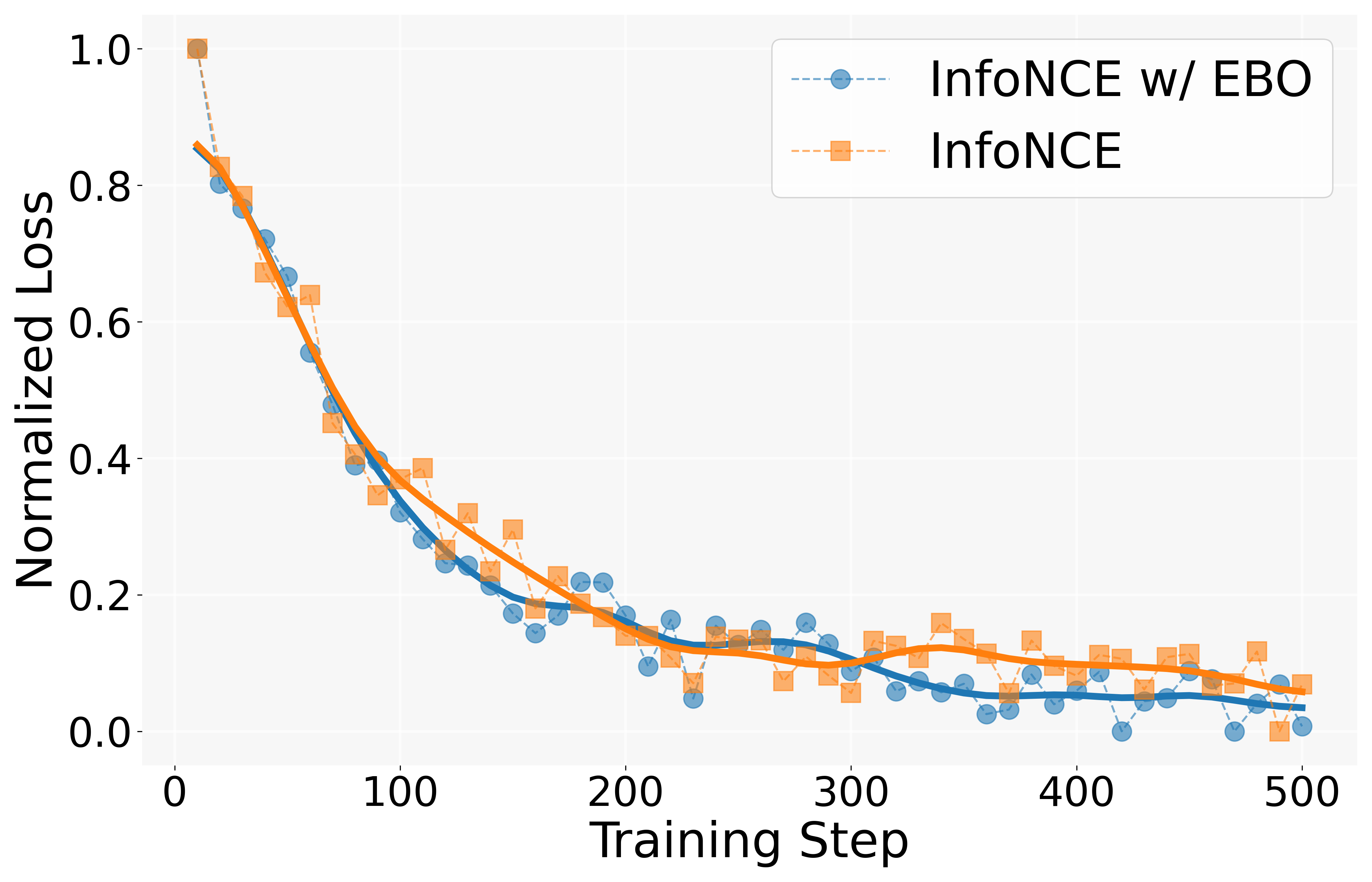}
\label{fig:loss_curve}
}
\caption{\textbf{Hyperparameter analysis.} Experiments are built upon LinCIR+MoTa-Adapter, using CLIP-G as the backbone.}
\label{fig:hyperparams}
\end{figure*}

\subsection{Qualitative Results}
Figure~\ref{fig:vis} presents qualitative predictions from the baseline LinCIR~\cite{LinCIR} as well as the results after integrating the MoTa-Adapter. which further support the effectiveness of our method. MoTa-Adapter is capable of handling fine-grained semantic relationships across various scenarios, such as object composition (row 2-3), attribute manipulation (row 4) and quantity change (row 1), in which LinCIR fails.
 
\subsection{Ablation Study} 
\textbf{Architectural Design.} Table~\ref{tbl:abl} summarizes several model variants. TPT, which only introduces a single set of learnable task prompts (equivalent to CoOp~\cite{coop}), improves performance, indicating its role in reducing task discrepancy. Adding the MoE mechanism (MoE-TPT) yields further gains, suggesting its effectiveness in capturing different types of modifications. While MDM also brings performance gains, its impact is smaller than that of MoE-TPT, implying that task discrepancy is the more dominant factor. Combining MoE-TPT with MDM leads to additional enhancements, highlighting the presence of two intertwined discrepancies that need to be addressed simultaneously. Finally, applying entropy-based optimization (EBO) achieves the overall best performance. Ablation studies based on SEARLE~\cite{serealcirco} are in Appendix B.2.

\textbf{Impact of Different LLMs.} We adopt different LLMs for text-anchored triplet construction, and the results, as shown in Table~\ref{tbl:llm}, are quite similar. Considering both performance and efficiency, we finally select LLaMA3-8B~\cite{llama3} as our model of choice. Additionally, applying data filtering leads to further performance gains, demonstrating the effectiveness of our proposed filtering strategy.

\textbf{Sensitivity Analysis.} Figure~\ref{fig:ablcurve} illustrates the impact of different hyperparameters, including the number of task prompts (NTPs), the number of experts (NEs), and the exponential smoothing hyperparameter $\beta$ in EBO. The performance generally shows an initial increase followed by a decline, with the optimal hyperparameter combination, i.e., $NTPs=8,NEs=4,\beta= 1$, identified.

\textbf{Data efficiency of MoTa-Adapter.} As shown in Figure~\ref{fig:quantity}, the performance reaches satisfactory levels with tens of thousands of samples, highlighting the data efficiency of our method. Further increases in data size result in marginal improvements; therefore, the full set of 360k generated triplets is released to support future research efforts.

\textbf{Visualization of Training Loss.} Figure~\ref{fig:loss_curve} visualizes the normalized training loss. After applying entropy-based optimization (EBO), the loss decreases more rapidly and stabilizes, maintaining a lower value towards the end, which contributes to improved model performance. 

\section{Discussion}
Although our method greatly enhances the performance of inversion-based methods, it still has certain limitations. (1) The first issue relates to generalization, to be specific, in the construction of text-anchored triplets as discussed in Section~\ref{subsec_gen_triplet}. We expand the real-world dataset CC3M-595k~\cite{llava} into a text-anchored triplet dataset, i.e., MoTa-CIR-360k. This enables our method to achieve strong performance on real-domain benchmarks, including CIRR, CIRCO, and GeneCIS. Although it also leads to a notable improvement on FashionIQ, our approach slightly underperforms RTD~\cite{rtd}, which is trained exclusively with textual supervision. Therefore, constructing multi-domain triplet datasets to improve generalization is a promising direction. (2) The generated relative captions tend to exhibit somewhat repetitive syntactic structures. It is worthwhile to explore whether supervised fine-tuning (SFT) of the LLM, or alternative strategies, can enhance the diversity of sentence structures and improve linguistic variation.

\section{Conclusion}
In this paper, we propose a lightweight post-hoc framework to improve inversion-based zero-shot CIR methods. Our framework includes MoTa-Adapter, a parameter-efficient tuning strategy, and a scalable text-anchored triplet construction pipeline. MoTa-Adapter adapts the dual encoder on the generated text-anchored triplets to reduce task and modality discrepancies by modulating the inversion network’s input on the image side and introducing learnable task prompts on the text side. These prompts are integrated via an MoE mechanism to capture task-specific priors and handle different modifications in the relative caption. Notably, to avoid the high computational cost of full fine-tuning, we are the first to introduce a lightweight adapter for zero-shot CIR, significantly enhancing performance while reducing training overhead. Experiments on four widely used benchmarks demonstrate the effectiveness of our approach.

\bibliography{aaai2026}

\end{document}